\documentclass[10pt,twocolumn,letterpaper]{article}

\usepackage{cvpr}
\usepackage{times}
\usepackage{epsfig}
\usepackage{graphicx}
\usepackage{amsmath}
\usepackage{amssymb}
\usepackage{tabularx}
\usepackage{subfigure}

\usepackage{verbatim}
\usepackage{authblk}

\usepackage{graphicx}
\usepackage{epstopdf}
\usepackage{subfigure}
\usepackage{footnote}
\newcommand{\tabincell}[2]{\begin{tabular}{@{}#1@{}}#2\end{tabular}}

\newcommand*\samethanks[1][\value{footnote}]{\footnotemark[#1]}
\usepackage[pagebackref=true,breaklinks=true,letterpaper=true,colorlinks,bookmarks=false]{hyperref}

\cvprfinalcopy 

\def\cvprPaperID{1797} 

\ifcvprfinal\fi
\begin{document}

\title{CosFace: Large Margin Cosine Loss for Deep Face Recognition}

\author{Hao Wang, Yitong Wang, Zheng Zhou, Xing Ji, Dihong Gong, Jingchao Zhou, \\
Zhifeng Li\thanks{Corresponding authors},{ } and Wei Liu\samethanks}
\affil{Tencent AI Lab \authorcr
{\tt\small \{hawelwang,yitongwang,encorezhou,denisji,sagazhou,michaelzfli\}@tencent.com }
{\tt\small 
gongdihong@gmail.com
wliu@ee.columbia.edu
}
}

\maketitle
\thispagestyle{empty}

\begin{abstract}
Face recognition has made extraordinary progress owing to the advancement of deep convolutional neural networks (CNNs). The central task of face recognition, including face verification and identification, involves face feature discrimination. However, the traditional softmax loss of deep CNNs usually lacks the power of discrimination. To address this problem, recently several loss functions such as center loss, large margin softmax loss, and angular softmax loss have been proposed. All these improved losses share the same idea: maximizing inter-class variance and minimizing intra-class variance. In this paper, we propose a novel loss function, namely large margin cosine loss (LMCL), to realize this idea from a different perspective. More specifically, we reformulate the softmax loss as a cosine loss by $L_2$ normalizing both features and weight vectors to remove radial variations, based on which a cosine margin term is introduced to further maximize the decision margin in the angular space. As a result, minimum intra-class variance and maximum inter-class variance are achieved by virtue of normalization and cosine decision margin maximization. We refer to our model trained with LMCL as CosFace. Extensive experimental evaluations are conducted on the most popular public-domain face recognition datasets such as MegaFace Challenge, Youtube Faces (YTF) and Labeled Face in the Wild (LFW). We achieve the state-of-the-art performance on these benchmarks, which confirms the effectiveness of our proposed approach.

\end{abstract}

\begin{figure}[t]
\begin{center}
   \includegraphics[width=1.0\linewidth, keepaspectratio]{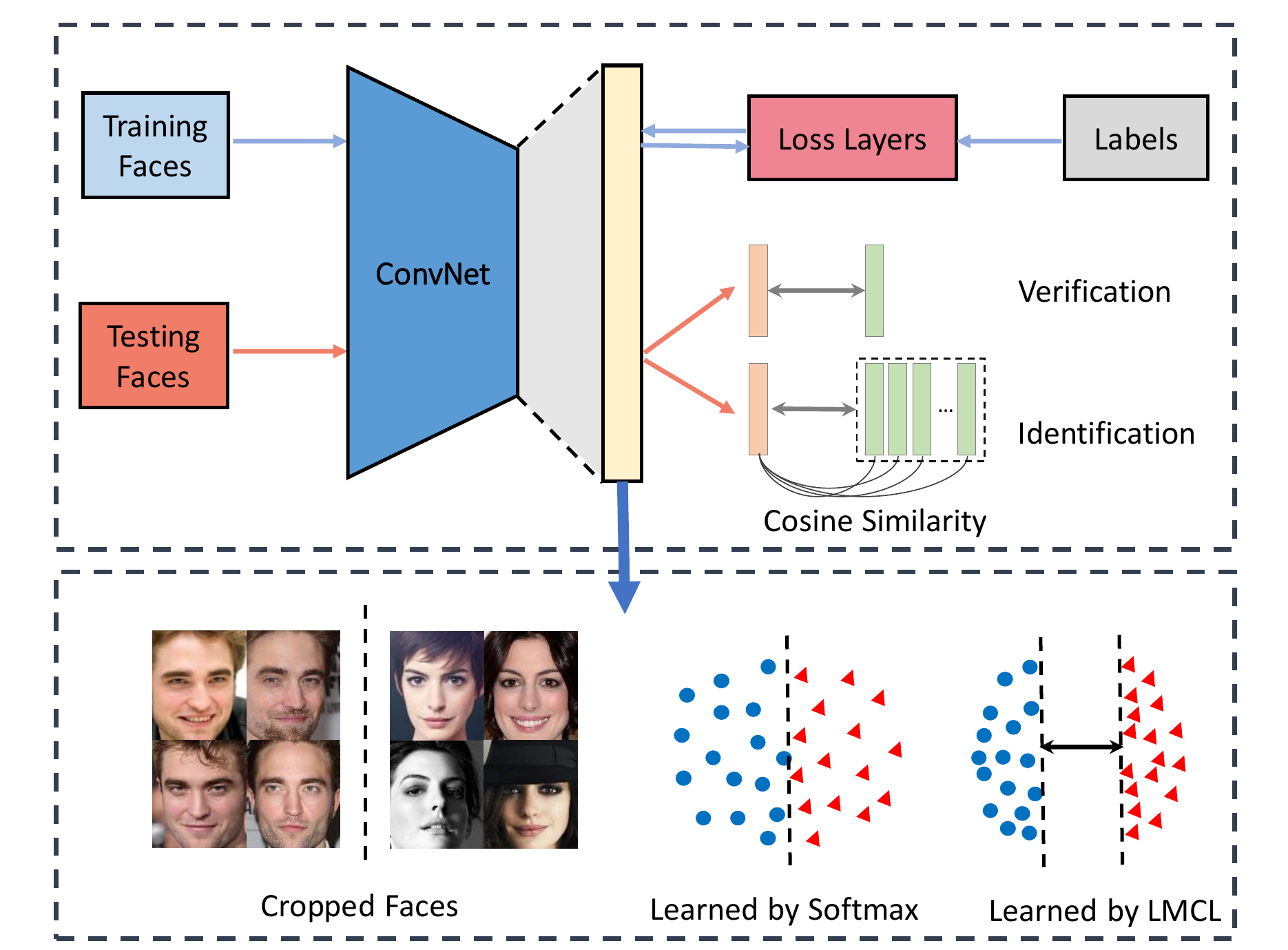}
\end{center}
   \caption{An overview of the proposed CosFace framework. In the training phase, the discriminative face features are learned with a large margin between different classes. In the testing phase, the testing data is fed into CosFace to extract face features which are later used to compute the cosine similarity score to perform face verification and identification. }
\label{fig:1}
\end{figure}

\section{Introduction}
Recently progress on the development of deep convolutional neural networks (CNNs) \cite{vgg,alexnet,senet,ResNet101,resnext} has significantly advanced the state-of-the-art performance on a wide variety of computer vision tasks, which makes deep CNN a dominant machine learning approach for computer vision. Face recognition, as one of the most common computer vision tasks, has been extensively studied for decades \cite{turk1991face,Xiong:2013:FRV:2586117.2586933,spatio-temporal,Li2009NonparametricDA,Li05nonparametricsubspace,Wang:2004:UFS:1018034.1018362,Belhumeur:1997:EVF:261506.261512}. Early studies build shallow models with low-level face features, while modern face recognition techniques are greatly advanced driven by deep CNNs.
Face recognition usually includes two sub-tasks: face verification and face identification. Both of these two tasks involve three stages: face detection, feature extraction, and classification. A deep CNN is able to extract clean high-level features, making itself possible to achieve superior performance with a relatively simple classification architecture: usually, a multilayer perceptron networks followed by a softmax loss \cite{deepface,deepid}. However, recent studies \cite{centerloss,lsoftmax,sphereface} found that the traditional softmax loss is insufficient to acquire the discriminating power for classification.

To encourage better discriminating performance, many research studies have been carried out \cite{centerloss,contrastiveloss1,contrastiveloss2,tripletloss1,tripletloss2,sphereface}. All these studies share the same idea for maximum discrimination capability: maximizing inter-class variance and minimizing intra-class variance. For example, \cite{centerloss,contrastiveloss1,contrastiveloss2,tripletloss1,tripletloss2} propose to adopt multi-loss learning in order to increase the feature discriminating power.
While these methods improve classification performance over the traditional softmax loss, they usually come with some extra limitations. For \cite{centerloss}, it only explicitly minimizes the intra-class variance while ignoring the inter-class variances, which may result in suboptimal solutions. \cite{contrastiveloss1,contrastiveloss2,tripletloss1,tripletloss2} require thoroughly scheming the mining of pair or triplet samples, which is an extremely time-consuming procedure. Very recently, \cite{sphereface} proposed to address this problem from a different perspective. More specifically, \cite{sphereface} (A-softmax) projects the original Euclidean space of features to an angular space, and introduces an angular margin for larger inter-class variance.

Compared to the Euclidean margin suggested by \cite{centerloss,contrastiveloss1,tripletloss1}, the angular margin is preferred because the cosine of the angle has intrinsic consistency with softmax. The formulation of cosine matches
the similarity measurement that is frequently applied to face recognition. From this perspective, it is more reasonable to directly introduce cosine margin between different classes to improve the cosine-related discriminative information.

 In this paper, we reformulate the softmax loss as a cosine loss by $L_2$ normalizing both features and weight vectors to remove radial variations, based on which a cosine margin term \emph{$m$} is introduced to further maximize the decision margin in the angular space. Specifically, we propose a novel algorithm, dubbed Large Margin Cosine Loss (LMCL), which takes the normalized features as input to learn highly discriminative features by maximizing the inter-class cosine margin. Formally, we define a hyper-parameter $m$ such that the decision boundary is given by $cos(\theta_1) - m  = cos(\theta_2)$, where $\theta_i$ is the angle between the feature and weight of class $i$.

For comparison, the decision boundary of the A-Softmax is defined over the angular space by $\cos(m\theta_1) = \cos(\theta_2)$, which has a difficulty in optimization due to the non-monotonicity of the cosine function. To overcome such a difficulty, one has to employ an extra trick with an ad-hoc piecewise function for A-Softmax. More importantly, the decision margin of A-softmax depends on $\theta$, which leads to different margins for different classes. As a result, in the decision space, some inter-class features have a larger margin while others have a smaller margin, which reduces the discriminating power. Unlike A-Softmax, our approach defines the decision margin in the cosine space, thus avoiding the aforementioned shortcomings.

Based on the LMCL, we build a sophisticated deep model called CosFace, as shown in Figure \ref{fig:1}. In the training phase, LMCL guides the ConvNet to learn features with a large cosine margin. In the testing phase, the face features are extracted from the ConvNet to perform either face verification or face identification. We summarize the contributions of this work as follows:

(1) We embrace the idea of maximizing inter-class variance and minimizing intra-class variance and propose a novel loss function, called LMCL, to learn highly discriminative deep features for face recognition.

(2) We provide reasonable theoretical analysis based on the hyperspherical feature distribution encouraged by LMCL.

(3) The proposed approach advances the state-of-the-art performance over most of the benchmarks on popular face databases including LFW\cite{lfw}, YTF\cite{ytface} and Megaface \cite{mf1,mf2}.

\section{Related Work}

\textbf{Deep Face Recognition.}
Recently, face recognition has achieved significant progress thanks to the great success of deep CNN models \cite{alexnet, vgg, googlenet, ResNet101}. In DeepFace \cite{deepface} and DeepID \cite{deepid}, face recognition is treated as a multi-class classification problem and deep CNN models are first introduced to learn features on large multi-identities datasets. DeepID2 \cite{deepid2} employs identification and verification signals to achieve better feature embedding. Recent works DeepID2+ \cite{deepid2plus} and DeepID3 \cite{deepid3} further explore the advanced network structures to boost recognition performance. FaceNet \cite{facenet} uses triplet loss to learn an Euclidean space embedding and a deep CNN is then trained on nearly 200 million face images, leading to the state-of-the-art performance. Other approaches \cite{fudan,hu2015face} also prove the effectiveness of deep CNNs on face recognition.

\textbf{Loss Functions.}
Loss function plays an important role in deep feature learning. Contrastive loss \cite{contrastiveloss1,contrastiveloss2} and triplet loss \cite{tripletloss1,tripletloss2} are usually used to increase the Euclidean margin for better feature embedding. Wen \emph{et al.} \cite{centerloss} proposed a center loss to learn centers for deep features of each identity and used the centers to reduce intra-class variance. Liu \emph{et al.} \cite{lsoftmax} proposed a large margin softmax (L-Softmax) by adding angular constraints to each identity to improve feature discrimination. Angular softmax (A-Softmax) \cite{sphereface} improves L-Softmax by normalizing the weights, which achieves better performance on a series of open-set face recognition benchmarks \cite{lfw,ytface,mf1}.
Other loss functions \cite{rangeloss,marginalloss,quadrupletloss,islandloss} based on contrastive loss or center loss also demonstrate the performance on enhancing discrimination.

\textbf{Normalization Approaches.}
Normalization has been studied in recent deep face recognition studies.
\cite{normface} normalizes the weights which replace the inner product with cosine similarity within the softmax loss. \cite{l2softmax} applies the $L_2$ constraint on features to embed faces in the normalized space. Note that normalization on feature vectors or weight vectors achieves much lower intra-class angular variability by concentrating more on the angle during training.  Hence the angles between identities can be well optimized. The von Mises-Fisher (vMF) based methods \cite{vmf,vmf_face} and A-Softmax \cite{sphereface} also adopt normalization in feature learning.


\section{Proposed Approach}
In  this section, we firstly introduce the proposed LMCL in detail (Sec. 3.1). And a comparison with other loss functions is given to show the superiority of the LMCL (Sec. 3.2). The feature normalization technique adopted by the LMCL is further described to clarify its effectiveness (Sec. 3.3). Lastly, we present a theoretical analysis for the proposed LMCL (Sec. 3.4).

\subsection{Large Margin Cosine Loss}
We start by rethinking the softmax loss from a cosine perspective. The softmax loss separates features from different classes by maximizing the posterior probability of the ground-truth class. Given an input feature vector $x_i$ with its corresponding label $y_i$, the softmax loss can be formulated as:
\begin{equation}\label{1}
 L_{s} =  \frac{1}{N}\sum_{i=1}^N{-\log{p_i}} = \frac{1}{N}\sum_{i=1}^N{-\log{\frac{e^{f_{y_i}}}{\sum_{j=1}^C{e^{f_j}}}}},
\end{equation}
where $p_i$ denotes the posterior probability of $x_i$ being correctly classified. $N$ is the number of training samples and $C$ is the number of classes. $f_j$ is usually denoted as activation of a fully-connected layer with weight vector $W_j$ and bias $B_j$.
We fix the bias $B_j=0$ for simplicity, and as a result $f_j$ is given by:
\begin{equation}\label{2}
 f_{j} = W_j^Tx = \Vert{W_j}\Vert \Vert{x}\Vert \cos{\theta_j},
\end{equation}
where $\theta_j$ is the angle between $W_j$ and $x$. This formula suggests that both norm and angle of vectors contribute to the posterior probability.

To develop effective feature learning, the norm of $W$ should be necessarily invariable. To this end, We fix $\Vert{W_j}\Vert = 1$ by $L_2$ normalization.
In the testing stage, the face recognition score of a testing face pair is usually calculated according to cosine similarity between the two feature vectors.
This suggests that the norm of feature vector $x$ is not contributing to the scoring function.
Thus, in the training stage, we fix $\Vert{x}\Vert=s$. Consequently, the posterior probability merely relies on cosine of angle.
The modified loss can be formulated as

\begin{equation}\label{1}
 L_{ns} = \frac{1}{N}\sum_{i}{-\log{\frac{e^{s \cos(\theta_{{y_i}, i})}}{\sum_{j}{e^{s \cos(\theta_{j, i})}}}}}.
\end{equation}
Because we remove variations in radial directions by fixing $\Vert{x}\Vert = s$, the resulting model learns features that are separable in the angular space. We refer to this loss as the Normalized version of Softmax Loss (NSL) in this paper.

However, features learned by the NSL are not sufficiently discriminative because the NSL only emphasizes correct classification.
To address this issue, we introduce the cosine margin to the classification boundary, which is naturally incorporated into the cosine formulation of Softmax.

Considering a scenario of binary-classes for example, let $\theta_{i}$ denote the angle between the learned feature vector and the weight vector of Class $C_i$ $(i=1,2)$.
The NSL forces $\cos(\theta_{1})>\cos(\theta_{2})$ for $C_1$, and similarly for $C_2$, so that features from different classes are correctly classified.
To develop a large margin classifier, we further require $\cos(\theta_{1}) - m >\cos(\theta_{2})$
and $\cos(\theta_{2}) - m >\cos(\theta_{1})$, where $m \geq 0$ is a fixed parameter
introduced to control the magnitude of the cosine margin.
Since $\cos(\theta_{i}) - m$ is lower than $\cos(\theta_{i})$,
the constraint is more stringent for classification.
The above analysis can be well generalized to the scenario of multi-classes.
Therefore, the altered loss reinforces the discrimination of learned features by encouraging an extra margin in the cosine space.

\begin{figure}
  \centering
    \includegraphics[width=0.5\textwidth, keepaspectratio]{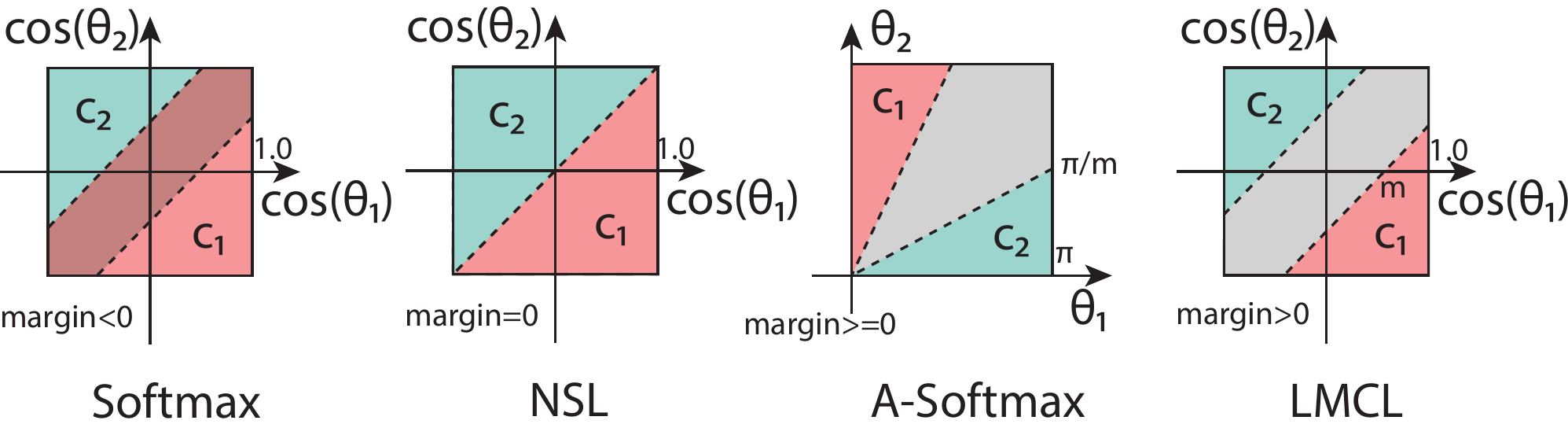}
  \caption{The comparison of decision margins for different loss functions the binary-classes scenarios. Dashed line represents decision boundary, and gray areas are decision margins.}
  \label{fig-margin-comp}
\end{figure}

Formally, we define the Large Margin Cosine Loss (LMCL) as:
\begin{equation}\label{1}
 L_{lmc} = \frac{1}{N}\sum_{i}{-\log{\frac{e^{s (\cos(\theta_{{y_i}, i}) - m)}}{e^{s (\cos(\theta_{{y_i}, i}) - m)} + \sum_{j \neq y_i}{e^{s \cos(\theta_{j, i})}}}}},
\end{equation}
subject to
\begin{equation}\label{1}
\begin{split}
  W &= \frac{W^*}{\Vert{W^*}\Vert}, \\
  x &= \frac{x^*}{\Vert{x^*}\Vert}, \\
  cos(\theta_j,i) &= {W_j}^Tx_i,
\end{split}
\end{equation}
where $N$ is the numer of training samples, $x_i$ is the $i$-th feature vector corresponding to the ground-truth class of $y_i$,
the $W_j$ is the weight vector of the $j$-th class, and $\theta_j$ is the angle between $W_j$ and $x_i$.

\subsection{Comparison on Different Loss Functions}
In this subsection, we compare the decision margin of our method (LMCL) to: Softmax, NSL, and A-Softmax, as illustrated in Figure~\ref{fig-margin-comp}.
For simplicity of analysis, we consider the binary-classes scenarios with classes $C_1$ and $C_2$. Let $W_1$ and $W_2$ denote weight vectors for $C_1$ and $C_2$, respectively.

\textbf{Softmax} loss defines a decision boundary by:
\begin{align*}
\Vert W_1\Vert \cos(\theta_1) = \Vert W_2\Vert \cos(\theta_2) .
\end{align*}
Thus, its boundary depends on both magnitudes of weight vectors and cosine of angles, which results in an overlapping decision area (margin $<$ 0) in the cosine space. This is illustrated in the first subplot of Figure~\ref{fig-margin-comp}. As noted before, in the testing stage it is a common strategy to only consider cosine similarity between testing feature vectors of faces. Consequently, the trained classifier with the Softmax loss is unable to perfectly classify testing samples in the cosine space.

\textbf{NSL} normalizes weight vectors $W_1$ and $W_2$ such that they have constant magnitude 1, which results in a decision boundary given by:
\begin{align*}
\cos(\theta_1) = \cos(\theta_2) .
\end{align*}
The decision boundary of NSL is illustrated in the second subplot of Figure~\ref{fig-margin-comp}. We can see that by removing radial variations, the NSL is able to perfectly classify testing samples in the cosine space, with margin = 0. However, it is not quite robust to noise because there is no decision margin: any small perturbation around the decision boundary can change the decision.

\textbf{A-Softmax} improves the softmax loss by introducing an extra margin, such that its decision boundary is given by:
\begin{align*}
C_1: \cos(m\theta_1) \ge \cos(\theta_2) ,\\
C_2: \cos(m\theta_2) \ge \cos(\theta_1) .
\end{align*}
Thus, for $C_1$ it requires $\theta_1 \le \frac{\theta_2}{m}$, and similarly for $C_2$. The third subplot of Figure~\ref{fig-margin-comp} depicts this decision area, where gray area denotes decision margin. However, the margin of A-Softmax is not consistent over all $\theta$ values: the margin becomes smaller as $\theta$ reduces, and vanishes completely when $\theta=0$. This results in two potential issues. First, for difficult classes $C_1$ and $C_2$ which are visually similar and thus have a smaller angle between $W_1$ and $W_2$, the margin is consequently smaller. Second, technically speaking one has to employ an extra trick with an ad-hoc piecewise function to overcome the nonmonotonicity difficulty of the cosine function.

\textbf{LMCL} (our proposed) defines a decision margin in cosine space rather than the angle space (like A-Softmax) by:
\begin{align*}
C_1: \cos(\theta_1) \ge \cos(\theta_2) + m , \\
C_2: \cos(\theta_2) \ge \cos(\theta_1) + m .
\end{align*}
Therefore, $\cos(\theta_1)$ is maximized while $\cos(\theta_2)$ being minimized for $C_1$ (similarly for $C_2$) to perform the large-margin classification.
The last subplot in Figure~\ref{fig-margin-comp} illustrates the decision boundary of LMCL in the cosine space, where we can see a clear margin($\sqrt{2}m$) in the produced distribution of the cosine of angle.
This suggests that the LMCL is more robust than the NSL, because a small perturbation around the decision boundary (dashed line) less likely leads to an incorrect decision.
The cosine margin is applied consistently to all samples, regardless of the angles of their weight vectors.


\subsection{Normalization on Features}
In the proposed LMCL, a normalization scheme is involved on purpose to derive the formulation of the cosine loss and remove variations in radial directions. Unlike \cite{sphereface} that only normalizes the weight vectors, our approach simultaneously normalizes both weight vectors and feature vectors. 
As a result, the feature vectors distribute on a hypersphere, where the scaling parameter $s$ controls the magnitude of radius. 
In this subsection, we discuss why feature normalization is necessary and how feature normalization encourages better feature learning in the proposed LMCL approach.

The necessity of feature normalization is presented in two respects:
First, the original softmax loss without feature normalization implicitly learns both the Euclidean norm ($L_2$-norm) of feature vectors and the cosine value of the angle. The $L_2$-norm is adaptively learned for minimizing the overall loss,
resulting in the relatively weak cosine constraint.
Particularly, the adaptive $L_2$-norm of easy samples becomes much larger than hard samples to remedy the inferior performance of cosine metric.
On the contrary, our approach requires the entire set of feature vectors to have the same $L_2$-norm
such that the learning only depends on cosine values to develop the discriminative power.
Feature vectors from the same classes are clustered together and those from different classes are pulled apart on the surface of the hypersphere.
Additionally, we consider the situation when the model initially starts to minimize the LMCL.
Given a feature vector $x$, let $\cos(\theta_i)$ and $\cos(\theta_j)$ denote cosine scores of the two classes, respectively.
Without normalization on features, the LMCL forces $\Vert x\Vert(\cos(\theta_i) - m) > \Vert x\Vert \cos(\theta_j)$.
Note that $\cos(\theta_i)$ and $\cos(\theta_j)$ can be initially comparable with each other.
Thus, as long as $(\cos(\theta_i) - m)$ is smaller than $\cos(\theta_j)$, $\Vert x\Vert$ is required to decrease for minimizing the loss, which degenerates the optimization.
Therefore, feature normalization is critical under the supervision of LMCL, especially when the networks are trained from scratch.
Likewise, it is more favorable to fix the scaling parameter $s$ instead of adaptively learning.

Furthermore, the scaling parameter $s$ should be set to a properly large value to yield better-performing features with lower training loss.
For NSL, the loss continuously goes down with higher $s$, while too small $s$ leads to an insufficient convergence even no convergence.
For LMCL, we also need adequately large $s$ to ensure a sufficient hyperspace for feature learning with an expected large margin.

In the following, we show the parameter $s$ should have a lower bound to obtain expected classification performance.
Given the normalized learned feature vector $x$ and unit weight vector $W$, we denote the total number of classes as $C$.
Suppose that the learned feature vectors separately lie on the surface of the hypersphere
and center around the corresponding weight vector. Let $P_W$ denote the expected minimum posterior probability of class center (\emph{i.e.}, $W$),
the lower bound of $s$ is given by \footnote{Proof is attached in the supplemental material.}:

\begin{equation}\label{1}
s \geq \frac{C-1}{C}\*\log{\frac{(C-1)\*P_W}{1-P_W}}.
\end{equation}

Based on this bound, we can infer that $s$ should be enlarged consistently
if we expect an optimal $P_w$ for classification with a certain number of classes.
Besides, by keeping a fixed $P_w$, the desired $s$
should be larger to deal with more classes since
the growing number of classes increase the difficulty for classification in the
relatively compact space. A hypersphere with large radius $s$ is therefore required for
embedding features with small intra-class distance and large inter-class distance.

\subsection{Theoretical Analysis for LMCL}

\begin{figure}
  \centering
    \includegraphics[width=0.4\textwidth, keepaspectratio]{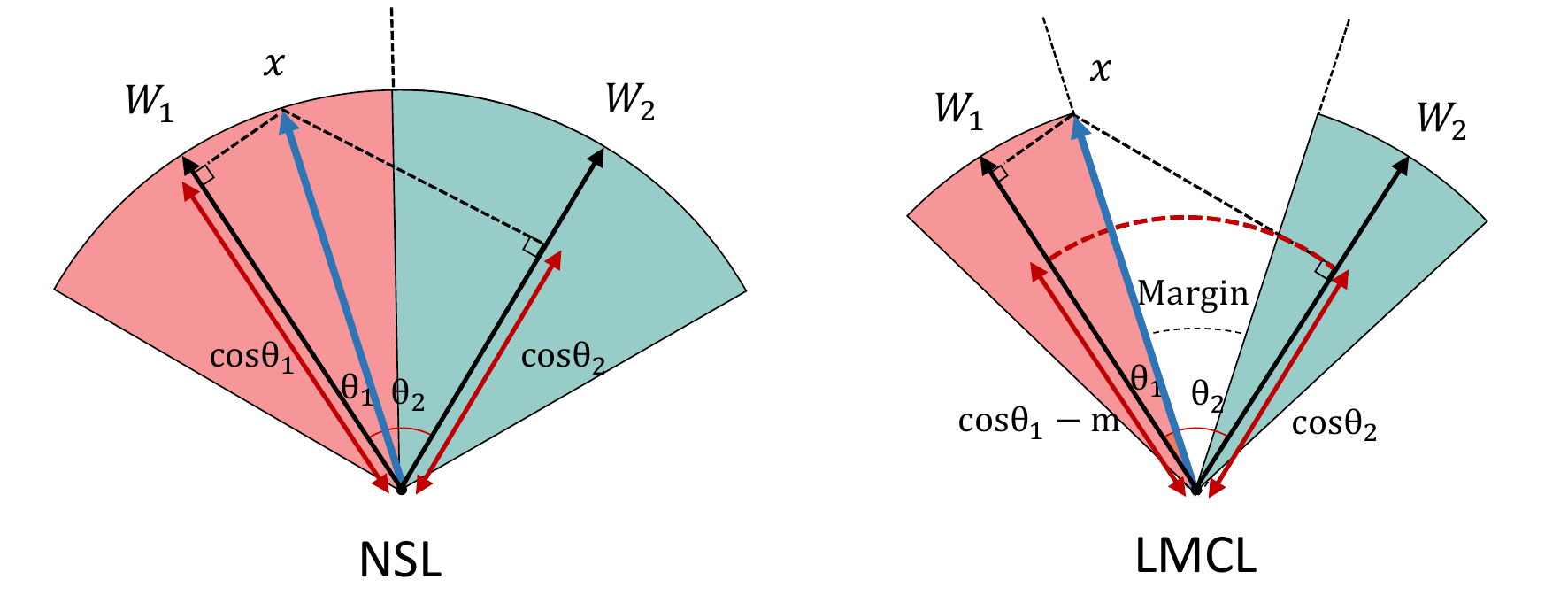}
  \caption{A geometrical interpretation of LMCL from feature perspective. Different color areas represent feature space from distinct classes. LMCL has a relatively compact feature region compared with NSL.}
  \label{fig-2d-angle}
\end{figure}

\begin{figure*}[t]
\begin{minipage}[b]{1.0\linewidth}
  \centering
  \centerline{\includegraphics[width=0.9\linewidth]{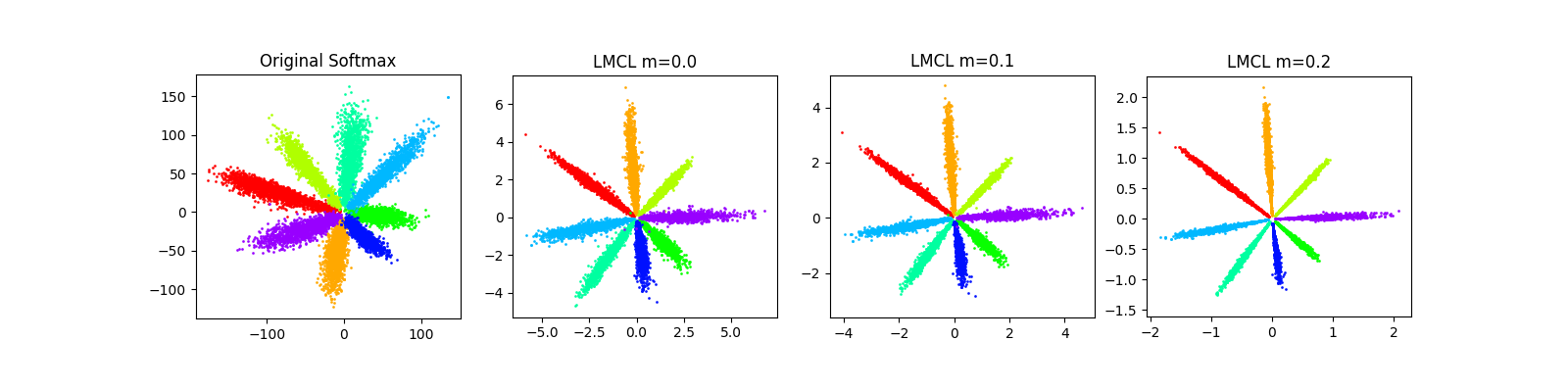}}
\end{minipage}
\begin{minipage}[b]{1.0\linewidth}
  \centering
  \centerline{\includegraphics[width=0.9\linewidth]{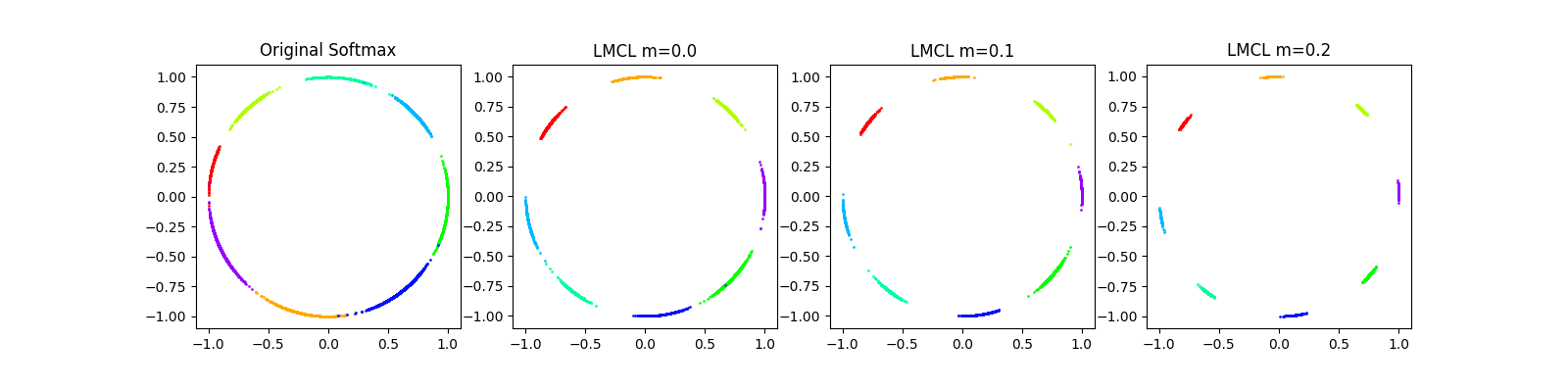}}
\end{minipage}
\caption{A toy experiment of different loss functions on 8 identities with 2D features. The first row maps the 2D features onto the Euclidean space, while the second row projects the 2D features onto the angular space. The gap becomes evident as the margin term $m$ increases.}
\label{fig:3}
\end{figure*}

The preceding subsections essentially discuss the LMCL from the classification point of view. In terms of learning the discriminative features on the hypersphere, the cosine margin servers as momentous part to strengthen the discriminating power of features. Detailed analysis about the quantitative feasible choice of the cosine margin (\emph{i.e.}, the bound of hyper-parameter $m$) is necessary. The optimal choice of $m$ potentially leads to more promising learning of highly discriminative face features. In the following, we delve into the decision boundary and angular margin in the feature space to derive the theoretical bound for hyper-parameter $m$.

First, considering the binary-classes case with classes $C_1$ and $C_2$ as before, suppose that the normalized feature vector $x$ is given.
Let $W_i$ denote the normalized weight vector, and $\theta_i$ denote the angle between $x$ and $W_i$.
For NSL, the decision boundary defines as $\cos\theta_1 - \cos\theta_2 = 0$, which is equivalent to the angular bisector of $W_1$ and $W_2$ as shown in the left of Figure \ref{fig-2d-angle}.
This addresses that the model supervised by NSL partitions the underlying feature space to two close regions, where the features near the boundary are extremely ambiguous (\emph{i.e.}, belonging to either class is acceptable).
In contrast, LMCL drives the decision boundary formulated by $\cos\theta_1 - \cos\theta_2 = m$ for $C_1$, in which $\theta_1$ should
be much smaller than $\theta_2$ (similarly for $C_2$). Consequently, the inter-class variance is enlarged while the intra-class variance shrinks.

Back to Figure~\ref{fig-2d-angle}, one can observe that the maximum angular margin is subject to the angle between $W_1$ and $W_2$.
Accordingly, the cosine margin should have the limited variable scope when $W_1$ and $W_2$ are given.
Specifically, suppose a scenario that all the feature vectors belonging to class $i$ exactly overlap with the corresponding weight vector $W_i$ of class $i$. In other words, every feature vector is identical to the weight vector for class $i$, and apparently the feature space is in an extreme situation, where all the feature vectors lie at their class center. In that case, the margin of decision boundaries has been maximized (\emph{i.e.}, the strict upper bound of the cosine margin).

To extend in general, we suppose that all the features are well-separated and we have a total number of $C$ classes. The theoretical variable scope of $m$ is supposed to be:
$0 \leq m \leq (1-\max(W_i^TW_j))$, where $i,j \leq n, i \neq j$.
The softmax loss tries to maximize the angle between any of the two weight vectors from two different classes in order to perform perfect classification. Hence, it is clear that the optimal solution for the softmax loss should uniformly distribute the weight vectors on a unit hypersphere. Based on this assumption, the variable scope of the introduced cosine margin $m$ can be inferred as follows \footnote{Proof is attached in the supplemental material.}:
\begin{equation}\label{1}
\begin{split}
& 0 \leq m \leq 1-\cos{\frac{2\pi}{C}},\quad  (K=2) \\
& 0 \leq m \leq \frac{C}{C-1}, \quad    (C \leq K+1)\\
& 0 \leq m \ll  \frac{C}{C-1}, \quad    (C > K+1)
\end{split}
\end{equation}
where $C$ is the number of training classes and $K$ is the dimension of learned features. The inequalities indicate that as the number of classes increases, the upper bound of the cosine margin between classes are decreased correspondingly.
Especially, if the number of classes is much larger than the feature dimension, the upper bound of the cosine margin will get even smaller.

A reasonable choice of larger $m \in [0, \frac{C}{C-1})$ should effectively boost the learning of highly discriminative features.
Nevertheless, parameter $m$ usually could not reach the theoretical upper bound in practice due to the vanishing of the feature space. That is, all the feature vectors are centered together according to the weight vector of the corresponding class.
In fact, the model fails to converge when $m$ is too large, because the cosine constraint (\emph{i.e.}, $\cos\theta_1 - m > \cos\theta_2$ or $\cos\theta_2 - m > \cos\theta_1$ for two classes) becomes stricter and is hard to be satisfied.
Besides, the cosine constraint with overlarge $m$ forces the training process to be more sensitive to noisy data. The ever-increasing $m$ starts to degrade the overall performance at some point because of failing to converge.

We perform a toy experiment for better visualizing on features and validating our approach. We select face images from 8 distinct identities containing enough samples to clearly show the feature points on the plot.
Several models are trained using the original softmax loss and the proposed LMCL with different settings of $m$.
We extract 2-D features of face images for simplicity. As discussed above, $m$ should be no larger than $1-\cos{\frac{\pi}{4}}$ (about 0.29), so we set up three choices of $m$ for comparison,
which are $m=0$, $m=0.1$, and $m=0.2$.
As shown in Figure \ref{fig:3}, the first row and second row present the feature distributions in Euclidean space and angular space, respectively.
We can observe that the original softmax loss produces ambiguity in decision boundaries while the proposed LMCL performs much better.
As $m$ increases, the angular margin between different classes has been amplified.

\section{Experiments}

\subsection{Implementation Details}

\textbf{Preprocessing.} Firstly, face area and landmarks are detected by MTCNN \cite{mtcnn} for the entire set of training and testing images.
Then, the 5 facial points (two eyes, nose and two mouth corners) are adopted to perform similarity transformation. After that
we obtain the cropped faces which are then resized to be $112\times 96$.
Following \cite{centerloss,sphereface}, each pixel (in [0, 255]) in RGB images is normalized by subtracting 127.5 then dividing by 128.

\textbf{Training.}
For a direct and fair comparison to the existing results that use small training datasets (less than 0.5M images and 20K subjects) \cite{mf1}, we train our models on a small training dataset, which is the publicly available CASIA-WebFace \cite{webface} dataset containing 0.49M face images from 10,575 subjects. 
We also use a large training dataset to evaluate the performance of our approach for benchmark comparison with the state-of-the-art results (using large training dataset) on the benchmark face dataset. The large training dataset that we use in this study is composed of several public datasets and a private face dataset, containing about
5M images from more than 90K identities. The training faces are horizontally flipped for data augmentation. In our experiments we remove face images belong to identities that appear in the testing datasets.

For the fair comparison, the CNN architecture used in our work is similar to \cite{sphereface}, which has 64 convolutional layers and is based on residual units\cite{ResNet101}. The scaling parameter $s$ in Equation (4) is set to 64 empirically.
We use Caffe\cite{caffe} to implement the modifications of the loss layer and run the models.
The CNN models are trained with SGD algorithm, with the batch size of 64 on 8 GPUs. The weight decay is set to 0.0005. 
For the case of training on the small dataset, the learning rate is initially 0.1 and divided by 10 at the 16K, 24K, 28k iterations, and we finish the training process at 30k iterations.
While the training on the large dataset terminates at 240k iterations, with the initial learning rate 0.05 dropped at 80K, 140K, 200K iterations.

\textbf{Testing.}
At testing stage,  features of original image and the flipped image are concatenated together to compose the final face representation.
The cosine distance of features is computed as the similarity score.
Finally, face verification and identification are conducted by thresholding and ranking the scores.
We test our models on several popular public face datasets, including LFW\cite{lfw}, YTF\cite{ytface}, and MegaFace\cite{mf1,mf2}.

\subsection{Exploratory Experiments}
\begin{figure}[t]
\begin{center}
   \includegraphics[width=0.8\linewidth, keepaspectratio]{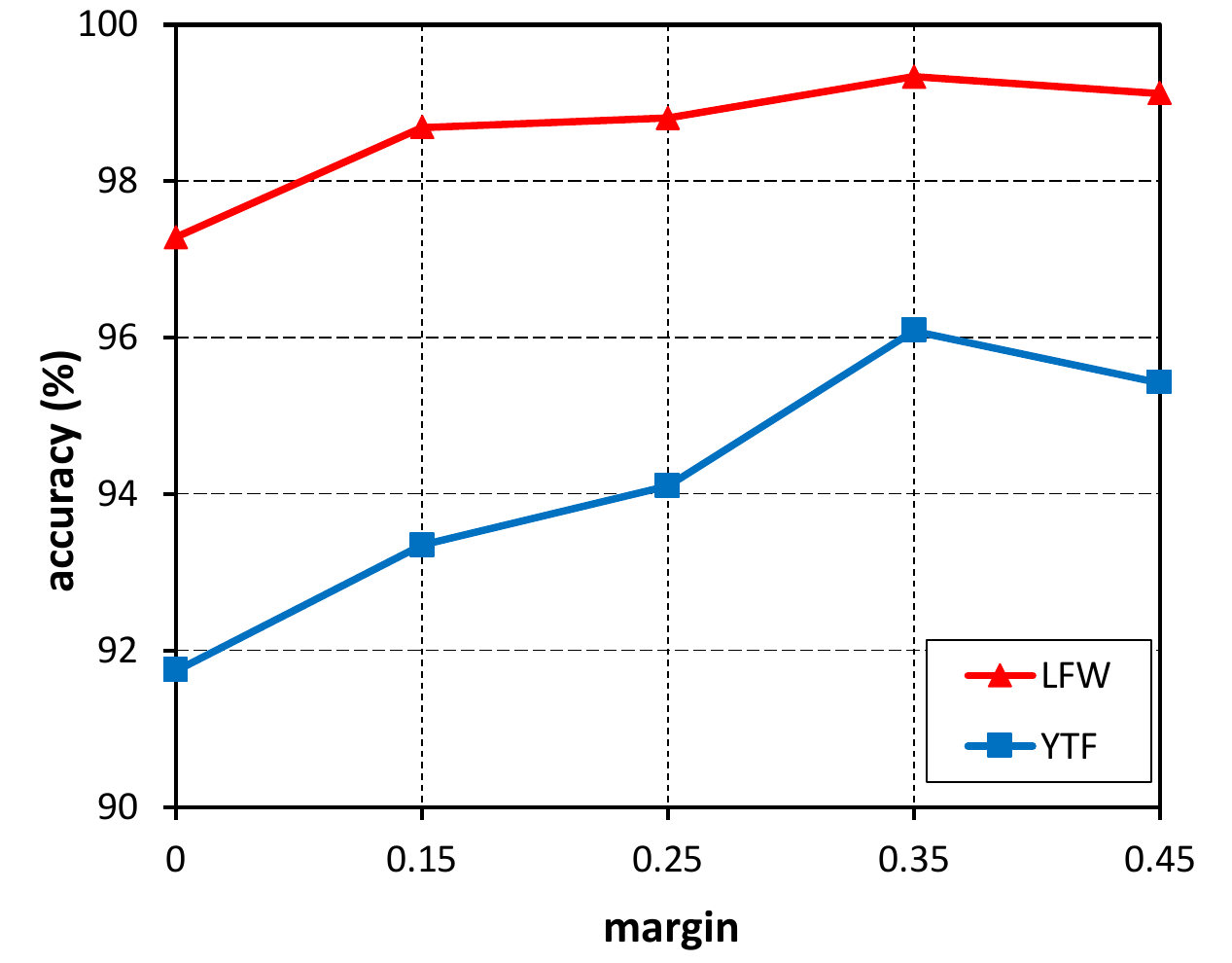}
\end{center}
   \caption{Accuracy (\%) of CosFace with different margin parameters $m$ on LFW\cite{lfw} and YTF \cite{ytface}.}
\label{fig:5}
\end{figure}

\textbf{Effect of $m$.}
The margin parameter $m$ plays a key role in LMCL. In this part we conduct an experiment to investigate the effect of $m$. By varying $m$ from 0 to 0.45 (If $m$ is larger than 0.45, the model will fail to converge), we use the small training data (CASIA-WebFace \cite{webface}) to train our CosFace model and evaluate its performance on the LFW\cite{lfw} and YTF\cite{ytface} datasets, as illustrated in Figure \ref{fig:5}.
We can see that the model without the margin (in this case m=0) leads to the worst performance.
As $m$ being increased, the accuracies are improved consistently on both datasets, and get saturated at $m=0.35$. This demonstrates the effectiveness of the margin $m$. By increasing the margin $m$, the discriminative power of the learned features can be significantly improved. In this study, $m$ is set to fixed 0.35 in the subsequent experiments.

\textbf{Effect of Feature Normalization.} To investigate the effect of the feature normalization scheme in our approach, we train our CosFace models on the CASIA-WebFace with and without the feature normalization scheme by fixing $m$ to 0.35, and compare their performance on LFW\cite{lfw}, YTF\cite{ytface}, and the Megaface Challenge 1(MF1)\cite{mf1}.
Note that the model trained without normalization is initialized by softmax loss and then supervised by the proposed LMCL. The comparative results are reported in Table \ref{norm-nonnorm}.
It is very clear that the model using the feature normalization scheme consistently outperforms the model without the feature normalization scheme across the three datasets.
As discussed above, feature normalization removes radical variance, and the learned features can be more discriminative in angular space. This experiment verifies this point.

\begin{table}
\footnotesize
\begin{center}
\begin{tabular}{|c|c|c|c|c|}
\hline
Normalization    & LFW      & YTF           & MF1 Rank 1	            & MF1 Veri. \\
\hline\hline
No               & 99.10    & 93.1          & 75.10                    & 88.65  \\
Yes              & 99.33    & 96.1          & 77.11                   & 89.88 \\
\hline

\end{tabular}
\end{center}
\caption{Comparison of our models with and without feature normalization on Megaface Challenge 1 (MF1). ``Rank 1" refers to rank-1 face identification accuracy and ``Veri." refers to face verification TAR (True Accepted Rate) under $10^{-6}$ FAR (False Accepted Rate).}
\label{norm-nonnorm}
\end{table}


\subsection{Comparison with state-of-the-art loss functions}

In this part, we compare the performance of the proposed LMCL with the state-of-the-art loss functions. Following the experimental setting in \cite{sphereface}, we train a model with the guidance of the proposed LMCL on the CAISA-WebFace\cite{webface}
using the same 64-layer CNN architecture described in \cite{sphereface}.
The experimental comparison on LFW, YTF and MF1 are reported in Table \ref{comprare_loss}.
For fair comparison, we are strictly following the model structure (a 64-layers ResNet-Like CNNs) and the detailed experimental settings of SphereFace \cite{sphereface}.
As can be seen in Table \ref{comprare_loss}, LMCL consistently
achieves competitive results compared to the other losses across the three datasets.
Especially, our method not only surpasses the performance of A-Softmax with feature normalization (named as A-Softmax-NormFea in Table \ref{comprare_loss}), but also significantly outperforms the other loss functions on YTF and MF1, which demonstrates the effectiveness of LMCL.

\begin{table}
\footnotesize
\begin{center}
\begin{tabular}{|c|c|c|c|c|c|c|}
\hline
Method 									& LFW 				& YTF             & \tabincell{c}{MF1\\Rank1}         & \tabincell{c}{MF1\\Veri.}     \\
\hline\hline
Softmax Loss \cite{sphereface}			                & 97.88 			& 93.1            & 54.85             & 65.92        \\
Softmax+Contrastive \cite{deepid2} 	 	& 98.78 			& 93.5            & 65.21             & 78.86         \\
Triplet	Loss \cite{facenet}				& 98.70 			& 93.4            & 64.79             & 78.32              \\
L-Softmax Loss \cite{lsoftmax}			& 99.10 			& 94.0            & 67.12             & 80.42               \\
Softmax+Center Loss \cite{centerloss}	& 99.05 			& 94.4	          & 65.49             & 80.14               \\
A-Softmax \cite{sphereface}				& \textbf{99.42} 	& 95.0            & 72.72             & 85.56              \\
A-Softmax-NormFea				& 99.32 	& 95.4            & 75.42             & 88.82              \\
\hline
\textbf{LMCL}						    & 99.33	            & \textbf{96.1}   & \textbf{77.11}    & \textbf{89.88}                \\
\hline
\end{tabular}
\end{center}
\caption{Comparison of the proposed LMCL with state-of-the-art loss functions in face recognition community. All the methods in this table are using the same training data and the same 64-layer CNN architecture.}
\label{comprare_loss}
\end{table}

\begin{table}
\footnotesize
\begin{center}
\begin{tabular}{|c|c|c|c|c|}
\hline
Method 								&Training Data  & \#Models	& LFW 				& YTF \\
\hline\hline
Deep Face\cite{deepface}				& 4M		& 3			& 97.35				& 91.4 \\
FaceNet\cite{facenet} 					& 200M 		& 1			& 99.63	            & 95.1 \\
DeepFR \cite{deepfr}					& 2.6M 		& 1			& 98.95 			& 97.3 \\
DeepID2+\cite{deepid2plus}				& 300K 		& 25		& 99.47 			& 93.2 \\
Center Face\cite{centerloss}			& 0.7M		& 1			& 99.28 			& 94.9 \\
Baidu\cite{baiduface}					& 1.3M 		& 1			& 99.13 			& - \\
SphereFace\cite{sphereface}				& 0.49M 	& 1			& 99.42 	        & 95.0 \\
\hline
\textbf{CosFace}				        & 5M 		& 1			& \textbf{99.73} 	& \textbf{97.6} \\
\hline
\end{tabular}
\end{center}
\caption{Face verification (\%) on the LFW and YTF datasets. ``\#Models" indicates the number of models that have been used in the method for evaluation.}
\label{lfw_ytf}
\end{table}

\begin{table}
\footnotesize
\begin{center}
\begin{tabular}{|c|c|c|c|}
\hline
Method &  Protocol & MF1 Rank1 & MF1 Veri. \\
\hline\hline
SIAT\underline{ }MMLAB\cite{centerloss}						& Small & 65.23 & 76.72 \\
DeepSense - Small 											& Small & 70.98 & 82.85 \\
SphereFace - Small\cite{sphereface} 						& Small & 75.76 & 90.04 \\
Beijing FaceAll V2							& Small & 76.66 & 77.60 \\
GRCCV 														& Small & 77.67 & 74.88 \\
FUDAN-CS\underline{ }SDS\cite{fudan} 	& Small & 77.98 & 79.19 \\
\hline
\textbf{CosFace(Single-patch)}             						    & Small & 77.11 & 89.88 \\
\textbf{CosFace(3-patch ensemble)}             						    & Small & \textbf{79.54} & \textbf{92.22} \\
\hline\hline
Beijing FaceAll\underline{ }Norm\underline{ }1600 & Large & 64.80 & 67.11 \\
Google - FaceNet v8\cite{facenet}								& Large & 70.49 & 86.47 \\
NTechLAB - facenx\underline{ }large 							& Large & 73.30 & 85.08 \\
SIATMMLAB TencentVision 										& Large & 74.20 & 87.27 \\
DeepSense V2 													& Large & 81.29 & 95.99 \\
YouTu Lab 														& Large & 83.29 & 91.34 \\
Vocord - deepVo V3 												& Large & \textbf{91.76} & 94.96  \\
\hline
\textbf{CosFace(Single-patch)}											& Large & 82.72 & 96.65     \\
\textbf{CosFace(3-patch ensemble)}              								& Large & 84.26 & \textbf{97.96}  \\
\hline
\end{tabular}
\end{center}
\caption{Face identification and verification evaluation on MF1. ``Rank 1" refers to rank-1 face identification accuracy and ``Veri." refers to face verification TAR under $10^{-6}$ FAR.}
\label{mf1}
\end{table}

\begin{table}
\footnotesize
\begin{center}
\begin{tabular}{|c|c|c|c|}
\hline
Method & Protocol & MF2 Rank1 & MF2 Veri. \\
\hline\hline
3DiVi			& Large & 57.04 & 66.45  \\
Team 2009							& Large & 58.93 & 71.12  \\
NEC 	& Large & 62.12 & 66.84  \\
GRCCV										& Large & 75.77 & 74.84  \\
SphereFace										& Large & 71.17 & 84.22  \\
\hline
\textbf{CosFace (Single-patch)}								& Large & 74.11 & 86.77\\
\textbf{CosFace(3-patch ensemble)}                 	& Large & \textbf{77.06} & \textbf{90.30}  \\
\hline
\end{tabular}
\end{center}
\caption{Face identification and verification evaluation on MF2. ``Rank 1" refers to rank-1 face identification accuracy and ``Veri." refers to face verification TAR under $10^{-6}$ FAR .}
\label{mf2}
\end{table}

\subsection{Overall Benchmark Comparison}


\subsubsection{Evaluation on LFW and YTF}
LFW \cite{lfw} is a standard face verification testing dataset in unconstrained conditions. It includes 13,233 face images from 5749 identities collected from the website. We evaluate our model strictly following the standard protocol of unrestricted with labeled outside data \cite{lfw}, and report the result on the 6,000 pair testing images.
YTF \cite{ytface} contains 3,425 videos of 1,595 different people. The average length of a video clip is 181.3 frames. All the video sequences were downloaded from YouTube. We follow the unrestricted with labeled outside data protocol and report the result on 5,000 video pairs.

As shown in Table \ref{lfw_ytf}, the proposed CosFace achieves state-of-the-art results of 99.73\% on LFW and 97.6\% on YTF. FaceNet achieves the runner-up performance on LFW with the large scale of the image dataset, which has approximately 200 million face images. In terms of YTF, our model reaches the first place over all other methods.

\subsubsection{Evaluation on MegaFace}
MegaFace \cite{mf1,mf2} is a very challenging testing benchmark recently released for large-scale face identification and verification, which contains a gallery set and a probe set. The gallery set in Megaface is composed of more than 1 million face images. 
The probe set has two existing databases: Facescrub \cite{facescrub} and FGNET \cite{fgnet}. 
In this study, we use the Facescrub dataset (containing 106,863 face images of 530 celebrities) as the probe set to evaluate the performance of our approach on both Megaface Challenge 1 and Challenge 2.

\textbf{MegaFace Challenge 1 (MF1).}
On the MegaFace Challenge 1 \cite{mf1}, The gallery set incorporates more than 1 million images from 690K individuals collected from Flickr photos \cite{flickr}.
Table \ref{mf1} summarizes the results of our models trained on two protocols of MegaFace where the training dataset is regarded as small if it has less than 0.5 million images, large otherwise. The CosFace approach shows its superiority for both the identification and verification tasks on both the protocols.



\textbf{MegaFace Challenge 2 (MF2).}
In terms of MegaFace Challenge 2 \cite{mf2}, all the algorithms need to use the training data provided by MegaFace. The training data for Megaface Challenge 2 contains 4.7 million faces and 672K identities, which corresponds to the large protocol. The gallery set has 1 million images that are different from the challenge 1 gallery set. 
Not surprisingly, Our method wins the first place of challenge 2 in table \ref{mf2}, setting a new state-of-the-art with a large margin (1.39\% on rank-1 identification accuracy and 5.46\% on verification performance).


\section{Conclusion}
In this paper, we proposed an innovative approach named LMCL to guide deep CNNs to learn highly discriminative face features. We provided a well-formed geometrical and theoretical interpretation to verify the effectiveness of the proposed LMCL. Our approach consistently achieves the state-of-the-art results on several face benchmarks. 
We wish that our substantial explorations on learning discriminative features via LMCL will benefit the face recognition community.

{\small
\bibliographystyle{ieee}
\bibliography{face}

\begin{thebibliography}{10}\itemsep=-1pt

\bibitem{fgnet}
{\em FG-NET Aging Database,http://www.fgnet.rsunit.com/}.

\bibitem{Belhumeur:1997:EVF:261506.261512}
P.~Belhumeur, J.~P. Hespanha, and D.~Kriegman.
\newblock Eigenfaces vs. fisherfaces: Recognition using class specific linear
  projection.
\newblock {\em IEEE Trans. Pattern Analysis and Machine Intelligence},
  19(7):711--720, July 1997.

\bibitem{islandloss}
J.~Cai, Z.~Meng, A.~S. Khan, Z.~Li, and Y.~Tong.
\newblock {Island Loss for Learning Discriminative Features in Facial
  Expression Recognition}.
\newblock {\em arXiv preprint arXiv:1710.03144}, 2017.

\bibitem{quadrupletloss}
W.~Chen, X.~Chen, J.~Zhang, and K.~Huang.
\newblock Beyond triplet loss: a deep quadruplet network for person
  re-identification.
\newblock {\em arXiv preprint arXiv:1704.01719}, 2017.

\bibitem{contrastiveloss1}
S.~Chopra, R.~Hadsell, and Y.~LeCun.
\newblock Learning a similarity metric discriminatively, with application to
  face verification.
\newblock In {\em {Conference on Computer Vision and Pattern Recognition
  (CVPR)}}, 2005.

\bibitem{marginalloss}
J.~Deng, Y.~Zhou, and S.~Zafeiriou.
\newblock Marginal loss for deep face recognition.
\newblock In {\em {Conference on Computer Vision and Pattern Recognition
  Workshops (CVPRW)}}, 2017.

\bibitem{contrastiveloss2}
R.~Hadsell, S.~Chopra, and Y.~LeCun.
\newblock Dimensionality reduction by learning an invariant mapping.
\newblock In {\em {Conference on Computer Vision and Pattern Recognition
  (CVPR)}}, 2006.

\bibitem{vmf_face}
M.~A. Hasnat, J.~Bohne, J.~Milgram, S.~Gentric, and L.~Chen.
\newblock {von Mises-Fisher Mixture Model-based Deep learning: Application to
  Face Verification}.
\newblock {\em arXiv preprint arXiv:1706.04264}, 2017.

\bibitem{ResNet101}
K.~He, X.~Zhang, S.~Ren, and J.~Sun.
\newblock {Deep Residual Learning for Image Recognition}.
\newblock In {\em {Conference on Computer Vision and Pattern Recognition
  (CVPR)}}, 2016.

\bibitem{tripletloss1}
E.~Hoffer and N.~Ailon.
\newblock Deep metric learning using triplet network.
\newblock In {\em International Workshop on Similarity-Based Pattern
  Recognition}, 2015.

\bibitem{hu2015face}
G.~Hu, Y.~Yang, D.~Yi, J.~Kittler, W.~Christmas, S.~Z. Li, and T.~Hospedales.
\newblock {When face recognition meets with deep learning: an evaluation of
  convolutional neural networks for face recognition}.
\newblock In {\em {International Conference on Computer Vision Workshops
  (ICCVW)}}, 2015.

\bibitem{senet}
J.~Hu, L.~Shen, and G.~Sun.
\newblock {Squeeze-and-Excitation Networks}.
\newblock {\em {arXiv preprint arXiv:1709.01507}}, 2017.

\bibitem{lfw}
G.~B. Huang, M.~Ramesh, T.~Berg, and E.~Learned-Miller.
\newblock Labeled faces in the wild: A database for studying face recognition
  in unconstrained environments.
\newblock In {\em Technical Report 07-49, University of Massachusetts,
  Amherst}, 2007.

\bibitem{caffe}
Y.~Jia, E.~Shelhamer, J.~Donahue, S.~Karayev, J.~Long, R.~Girshick,
  S.~Guadarrama, and T.~Darrell.
\newblock {Caffe: Convolutional architecture for fast feature embedding}.
\newblock In {\em {Proceedings of the 2016 ACM on Multimedia Conference (ACM
  MM)}}, 2014.

\bibitem{vgg}
{K. Simonyan and A. Zisserman}.
\newblock {Very deep convolutional networks for large-scale image recognition}.
\newblock In {\em {International Conference on Learning Representations
  (ICLR)}}, 2015.

\bibitem{mtcnn}
{K. Zhang, Z. Zhang, Z. Li and Y. Qiao}.
\newblock {Joint Face Detection and Alignment using Multi-task Cascaded
  Convolutional Networks}.
\newblock {\em {Signal Processing Letters}}, 23(10):1499--1503, 2016.

\bibitem{mf1}
I.~Kemelmacher-Shlizerman, S.~M. Seitz, D.~Miller, and E.~Brossard.
\newblock The megaface benchmark: 1 million faces for recognition at scale.
\newblock In {\em {Conference on Computer Vision and Pattern Recognition
  (CVPR)}}, 2016.

\bibitem{alexnet}
A.~Krizhevsky, I.~Sutskever, and G.~E. Hinton.
\newblock {Imagenet classification with deep convolutional neural networks}.
\newblock In {\em {Advances in Neural Information Processing Systems (NIPS)}},
  2012.

\bibitem{Li2009NonparametricDA}
Z.~Li, D.~Lin, and X.~Tang.
\newblock Nonparametric discriminant analysis for face recognition.
\newblock {\em IEEE Transactions on Pattern Analysis and Machine Intelligence},
  31:755--761, 2009.

\bibitem{Li05nonparametricsubspace}
Z.~Li, W.~Liu, D.~Lin, and X.~Tang.
\newblock Nonparametric subspace analysis for face recognition.
\newblock In {\em Conference on Computer Vision and Pattern Recognition
  (CVPR)}, 2005.

\bibitem{baiduface}
J.~Liu, Y.~Deng, T.~Bai, Z.~Wei, and C.~Huang.
\newblock Targeting ultimate accuracy: Face recognition via deep embedding.
\newblock {\em arXiv preprint arXiv:1506.07310}, 2015.

\bibitem{spatio-temporal}
W.~Liu, Z.~Li, and X.~Tang.
\newblock Spatio-temporal embedding for statistical face recognition from
  video.
\newblock In {\em European Conference on Computer Vision (ECCV)}, 2006.

\bibitem{sphereface}
W.~Liu, Y.~Wen, Z.~Yu, M.~Li, B.~Raj, and L.~Song.
\newblock {SphereFace: Deep Hypersphere Embedding for Face Recognition}.
\newblock In {\em {Conference on Computer Vision and Pattern Recognition
  (CVPR)}}, 2017.

\bibitem{lsoftmax}
W.~Liu, Y.~Wen, Z.~Yu, and M.~Yang.
\newblock {Large-Margin Softmax Loss for Convolutional Neural Networks}.
\newblock In {\em {International Conference on Machine Learning (ICML)}}, 2016.

\bibitem{mf2}
A.~Nech and I.~Kemelmacher-Shlizerman.
\newblock Level playing field for million scale face recognition.
\newblock In {\em {Conference on Computer Vision and Pattern Recognition
  (CVPR)}}, 2017.

\bibitem{facescrub}
H.-W. Ng and S.~Winkler.
\newblock A data-driven approach to cleaning large face datasets.
\newblock In {\em Image Processing (ICIP), 2014 IEEE International Conference
  on}, pages 343--347. IEEE, 2014.

\bibitem{deepfr}
O.~M. Parkhi, A.~Vedaldi, A.~Zisserman, et~al.
\newblock Deep face recognition.
\newblock In {\em BMVC}, volume~1, page~6, 2015.

\bibitem{l2softmax}
R.~Ranjan, C.~D. Castillo, and R.~Chellappa.
\newblock {L2-constrained Softmax Loss for Discriminative Face Verification}.
\newblock {\em arXiv preprint arXiv:1703.09507}, 2017.

\bibitem{facenet}
F.~Schroff, D.~Kalenichenko, and J.~Philbin.
\newblock {Facenet: A unified embedding for face recognition and clustering}.
\newblock In {\em {Conference on Computer Vision and Pattern Recognition
  (CVPR)}}, 2015.

\bibitem{deepid2}
Y.~Sun, Y.~Chen, X.~Wang, and X.~Tang.
\newblock {Deep learning face representation by joint
  identification-verification}.
\newblock In {\em {Advances in Neural Information Processing Systems (NIPS)}},
  2014.

\bibitem{deepid3}
Y.~Sun, D.~Liang, X.~Wang, and X.~Tang.
\newblock {DeepID3: Face recognition with very deep neural networks}.
\newblock {\em arXiv preprint arXiv:1502.00873}, 2015.

\bibitem{deepid}
Y.~Sun, X.~Wang, and X.~Tang.
\newblock {Deep learning face representation from predicting 10,000 classes}.
\newblock In {\em {Conference on Computer Vision and Pattern Recognition
  (CVPR)}}, 2014.

\bibitem{deepid2plus}
Y.~Sun, X.~Wang, and X.~Tang.
\newblock {Deeply learned face representations are sparse, selective, and
  robust}.
\newblock In {\em {Conference on Computer Vision and Pattern Recognition
  (CVPR)}}, 2015.

\bibitem{googlenet}
C.~Szegedy, W.~Liu, Y.~Jia, P.~Sermanet, S.~Reed, D.~Anguelov, D.~Erhan,
  V.~Vanhoucke, and A.~Rabinovich.
\newblock Going deeper with convolutions.
\newblock In {\em {Conference on Computer Vision and Pattern Recognition
  (CVPR)}}, 2015.

\bibitem{deepface}
Y.~Taigman, M.~Yang, M.~Ranzato, and L.~Wolf.
\newblock {Deepface: Closing the gap to human-level performance in face
  verification}.
\newblock In {\em {Conference on Computer Vision and Pattern Recognition
  (CVPR)}}, 2014.

\bibitem{flickr}
B.~Thomee, D.~A. Shamma, G.~Friedland, B.~Elizalde, K.~Ni, D.~Poland, D.~Borth,
  and L.-J. Li.
\newblock {YFCC100M: The new data in multimedia research}.
\newblock {\em Communications of the ACM}, 2016.

\bibitem{turk1991face}
M.~A. Turk and A.~P. Pentland.
\newblock Face recognition using eigenfaces.
\newblock In {\em {Conference on Computer Vision and Pattern Recognition
  (CVPR)}}, 1991.

\bibitem{normface}
F.~Wang, X.~Xiang, J.~Cheng, and A.~L. Yuille.
\newblock {NormFace: $ L\_2 $ Hypersphere Embedding for Face Verification}.
\newblock In {\em {Proceedings of the 2017 ACM on Multimedia Conference (ACM
  MM)}}, 2017.

\bibitem{tripletloss2}
J.~Wang, Y.~Song, T.~Leung, C.~Rosenberg, J.~Wang, J.~Philbin, B.~Chen, and
  Y.~Wu.
\newblock Learning fine-grained image similarity with deep ranking.
\newblock In {\em {Conference on Computer Vision and Pattern Recognition
  (CVPR)}}, 2014.

\bibitem{Wang:2004:UFS:1018034.1018362}
X.~Wang and X.~Tang.
\newblock A unified framework for subspace face recognition.
\newblock {\em IEEE Trans. Pattern Analysis and Machine Intelligence},
  26(9):1222--1228, Sept. 2004.

\bibitem{fudan}
Z.~Wang, K.~He, Y.~Fu, R.~Feng, Y.-G. Jiang, and X.~Xue.
\newblock {Multi-task Deep Neural Network for Joint Face Recognition and Facial
  Attribute Prediction}.
\newblock In {\em Proceedings of the 2017 ACM on International Conference on
  Multimedia Retrieval (ICMR)}, 2017.

\bibitem{centerloss}
Y.~Wen, K.~Zhang, Z.~Li, and Y.~Qiao.
\newblock A discriminative feature learning approach for deep face recognition.
\newblock In {\em European Conference on Computer Vision (ECCV)}, pages
  499--515, 2016.

\bibitem{ytface}
L.~Wolf, T.~Hassner, and I.~Maoz.
\newblock Face recognition in unconstrained videos with matched background
  similarity.
\newblock In {\em {Conference on Computer Vision and Pattern Recognition
  (CVPR)}}, 2011.

\bibitem{resnext}
S.~Xie, R.~Girshick, P.~Doll{\'a}r, Z.~Tu, and K.~He.
\newblock Aggregated residual transformations for deep neural networks.
\newblock {\em arXiv preprint arXiv:1611.05431}, 2016.

\bibitem{Xiong:2013:FRV:2586117.2586933}
Y.~Xiong, W.~Liu, D.~Zhao, and X.~Tang.
\newblock Face recognition via archetype hull ranking.
\newblock In {\em IEEE International Conference on Computer Vision (ICCV)},
  2013.

\bibitem{webface}
D.~Yi, Z.~Lei, S.~Liao, and S.~Z. Li.
\newblock Learning face representation from scratch.
\newblock {\em arXiv preprint arXiv:1411.7923}, 2014.

\bibitem{rangeloss}
X.~Zhang, Z.~Fang, Y.~Wen, Z.~Li, and Y.~Qiao.
\newblock {Range Loss for Deep Face Recognition with Long-tail}.
\newblock In {\em {International Conference on Computer Vision (ICCV)}}, 2017.

\bibitem{vmf}
X.~Zhe, S.~Chen, and H.~Yan.
\newblock {Directional Statistics-based Deep Metric Learning for Image
  Classification and Retrieval}.
\newblock {\em arXiv preprint arXiv:1802.09662}, 2018.

\end{thebibliography}
}

\clearpage
\appendix
\section{Supplementary Material}
This supplementary document provides mathematical details for the derivation of the lower bound of the scaling parameter $s$ (Equation 6 in the main paper), and the variable scope of the cosine margin $m$ (Equation 7 in the main paper).\\


\begin{large}
\noindent
\textbf{Proposition of the Scaling Parameter $s$}
\end{large}
\vspace{4pt}

Given the normalized learned features $x$ and unit weight vectors $W$, we denote the total number of classes as $C$ where $C>1$.
Suppose that the learned features separately lie on the surface of a hypersphere
and center around the corresponding weight vector. Let $P_w$ denote the expected minimum posterior probability of the class center (\emph{i.e.}, $W$).
The lower bound of $s$ is formulated as follows:
\begin{align*}
s \geq \frac{C-1}{C}\*\ln{\frac{(C-1)\*P_W}{1-P_W}}
\end{align*}

\textbf{Proof:}

Let $W_i$ denote the $i$-th unit weight vector.
$\forall i$, we have:
\begin{equation}\label{8}\tag{8}
\frac{e^{s}}{e^{s} + \sum_{j, j \neq i}{e^{s({W_i^T}{W_j})}}} \geq P_{W},
\end{equation}

\begin{equation}\label{9}\tag{9}
1 + e^{-s}\*\sum_{j, j \neq i}{e^{s\*({W_i^T}{W_j})}} \leq \frac{1}{P_W},
\end{equation}

\begin{equation}\label{1}\tag{10}
\sum_{i=1}^C{(1 + e^{-s}\*\sum_{j, j \neq i}{e^{s({W_i^T}{W_j})}})} \leq \frac{C}{P_W},
\end{equation}

\begin{equation}\label{1}\tag{11}
1 + \frac{e^{-s}\*}{C}\sum_{i,j, i \neq j}{e^{s({W_i^T}{W_j})}} \leq \frac{1}{P_W}.
\end{equation}

Because $f(x)=e^{s\cdot x}$ is a convex function, according to Jensen's inequality, we obtain:
\begin{equation}\label{1}\tag{12}
\frac{1}{C(C-1)}\sum_{i,j, i \neq j}{e^{s({W_i^T}{W_j})}} \geq e^{\frac{s}{C(C-1)}{\sum_{i,j, i \neq j}{{W_i^T}{W_j}}}}.
\end{equation}
Besides, it is known that

\begin{equation}\label{1}\tag{13}
\sum_{i,j, i \neq j}{{W_i^T}{W_j}} = (\sum_{i}{W_i})^2 - (\sum_{i}{W_i^2}) \geq -C.
\end{equation}

Thus, we have:
\begin{equation}\label{1}\tag{14}
1 + {(C-1)}{e^{-\frac{s\*C}{C-1}}} \leq \frac{1}{P_W}.
\end{equation}
Further simplification yields:
\begin{equation}\label{1}\tag{15}
s \geq \frac{C-1}{C}\*\ln{\frac{(C-1)\*P_W}{1-P_W}}.
\end{equation}

The equality holds if and only if every ${W_i^T}{W_j}$ is equal ($i\ne j$), and $\sum_{i}{W_i} = 0$.
Because at most $K+1$ unit vectors are able to satisfy this condition in the K-dimension hyper-space, the equality holds only when $C \leq K+1$, where K is the dimension of the learned features.\\




\begin{large}
\noindent
\textbf{Proposition of the Cosine Margin $m$}
\end{large}
\vspace{4pt}

Suppose that the weight vectors are uniformly distributed on a unit hypersphere. 
The variable scope of the introduced cosine margin $m$ is formulated as follows :
\begin{align*}
& 0 \leq m \leq 1-\cos{\frac{2\pi}{C}},\quad  (K=2) \\
& 0 \leq m \leq \frac{C}{C-1}, \quad    (K>2, C \leq K+1)\\
& 0 \leq m \ll  \frac{C}{C-1}, \quad    (K>2, C > K+1)
\end{align*}
where $C$ is the total number of training classes and $K$ is the dimension of the learned features.

\textbf{Proof:}

For $K=2$, the weight vectors uniformly spread on a unit circle.
Hence, $\max(W_i^TW_j) = \cos{\frac{2\pi}{C}}$. It follows $0 \leq m \leq (1-\max(W_i^TW_j)) = 1-\cos{\frac{2\pi}{C}}$.

For $K > 2$, the inequality below holds:

\begin{align}\label{1}\tag{16}
C(C-1)\max(W_i^T\*W_j) 
&\geq \sum_{i,j, i \neq j}{{W_i^T}{W_j}}\nonumber\\
&= (\sum_{i}{W_i})^2 - (\sum_{i}{W_i^2})\nonumber\\
&\geq -C. \nonumber
\end{align}

Therefore, $\max(W_i^TW_j) \geq \frac{-1}{C-1}$, and we have $0 \leq m \leq (1-\max(W_i^T\*W_j)) \leq \frac{C}{C-1}$.

Similarly, the equality holds if and only if every ${W_i^T}{W_j}$ is equal ($i \neq j$), and $\sum_{i}{W_i} = 0$. As discussed above, this is satisfied only if $C \leq K+1$. On this condition, the distance between the vertexes of two arbitrary $W$ should be the same. In other words, they form a regular simplex such as an equilateral triangle if $C=3$, or a regular tetrahedron if $C=4$.

For the case of $C > K+1$, the equality cannot be satisfied. In fact, it is unable to formulate the strict upper bound. Hence, we obtain $0 \leq m \ll  \frac{C}{C-1}$.
Because the number of classes can be much larger than the feature dimension,
the equality cannot hold in practice.

\end{document}